\documentclass[journal]{IEEEtran}

\usepackage[english]{babel}
\usepackage[utf8x]{inputenc}
\usepackage[T1]{fontenc}

\usepackage{amsmath,amssymb,mathrsfs,bbm}
\usepackage{graphicx}
\usepackage[colorinlistoftodos]{todonotes}
\usepackage[colorlinks=true, allcolors=blue]{hyperref}
\usepackage{multirow}
\usepackage[lined,linesnumbered,ruled]{algorithm2e}

\usepackage{booktabs}
\usepackage{graphicx}
\usepackage{caption}
\usepackage{subcaption}
\usepackage[position=b]{subcaption}

\usepackage[noadjust]{cite}
\usepackage{multirow}
\usepackage[lined,linesnumbered,ruled]{algorithm2e}

\usepackage{psfrag,epsfig,graphics}
\usepackage{amsmath,amsthm,amssymb,multirow}
\usepackage{mathbbol}
\usepackage{amssymb}

\DeclareSymbolFontAlphabet{\amsmathbb}{AMSb}%

\newcommand{\cp}[1]{\ifmmode {\mathcal{#1}}\else ${\mathcal{#1}}$\fi}
	
\newcommand{\bA}{\boldsymbol{A}}
\newcommand{\bB}{\boldsymbol{B}}

\newcommand{\bI}{\boldsymbol{I}}

\newcommand{\bM}{\boldsymbol{M}}

\newcommand{\bX}{\boldsymbol{X}}
\newcommand{\bY}{\boldsymbol{Y}}
\newcommand{\bZ}{\boldsymbol{Z}}

\newcommand{\ba}{\boldsymbol{a}}

\newcommand{\bm}{\boldsymbol{m}}

\newcommand{\bp}{\boldsymbol{p}}
\newcommand{\be}{\boldsymbol{e}}

\newcommand{\by}{\boldsymbol{y}}
\newcommand{\bs}{\boldsymbol{s}}
\newcommand{\bu}{\boldsymbol{u}}

\newcommand{\bx}{\boldsymbol{x}}

\newcommand{\bz}{\boldsymbol{z}}

\newcommand{\calD}{\mathcal{D}}

\newcommand{\calG}{\mathcal{G}}

\newcommand{\bbM}{\mathbb{M}}

\newcommand{\bbX}{\mathbb{X}}

\newcommand{\bbZ}{\mathbb{Z}}

\newcommand{\bbPsi}{\mathbb{\Psi}}

\newcommand{\btheta}{\boldsymbol{\theta}}

\newcommand{\cb}[1]{\boldsymbol{#1}}

\newcommand{\Ex}{\operatorname{E}}

\usepackage{color}  

\definecolor{darkgreen}{rgb}{0., 0.4, 0.}

\newcommand\scalemath[2]{\scalebox{#1}{\mbox{\ensuremath{\displaystyle #2}}}}

\title{Deep Generative Endmember Modeling:\\ An Application to Unsupervised Spectral Unmixing}

\author{Ricardo~Augusto~Borsoi~\IEEEmembership{Student~Member,~IEEE}, Tales Imbiriba,~\IEEEmembership{Member,~IEEE,} Jos\'e~Carlos~Moreira~Bermudez,~\IEEEmembership{Senior~Member,~IEEE}
\thanks{This work has been supported by the National Council for Scientific and Technological Development (CNPq) under grants 304250/2017-1, 409044/2018-0, 141271/2017-5 and 204991/2018-8, and by the Brazilian Education Ministry (CAPES) under grant PNPD/1811213.}
\thanks{R.A. Borsoi is with the Department of Electrical Engineering, Federal University of Santa Catarina (DEE--UFSC), Florian\'opolis, SC, Brazil, and with the Lagrange Laboratory, Universit\'e  C\^ote  d'Azur, Nice, France. e-mail: \mbox{raborsoi@gmail.com}.
T. Imbiriba was with the DEE--UFSC, Florian\'opolis, SC, Brazil, and is with the ECE department of the Northeastern University, Boston, MA, USA. e-mail: \mbox{talesim@ece.neu.edu}. 
J.C.M. Bermudez is with the DEE--UFSC, Florian\'opolis, SC, Brazil, and with the Graduate Program on Electronic Engineering and Computing, Catholic University of Pelotas (UCPel) Pelotas, Brazil. e-mail: \mbox{j.bermudez@ieee.org}.}
\thanks{Manuscript received Month day, year; revised Month day, year.}}

\begin{document}
\maketitle

\begin{abstract}
Endmember (EM) spectral variability can greatly impact the performance of standard hyperspectral image analysis algorithms.
Extended parametric models have been successfully applied to account for the EM spectral variability. However, these models still lack the compromise between flexibility and low-dimensional representation that is necessary to properly explore the fact that spectral variability is often confined to a low-dimensional manifold in real scenes.  
In this paper we propose to learn a spectral variability model directly from the observed data, instead of imposing it \emph{a priori}. This is achieved through a deep generative EM model, which is estimated using a variational autoencoder (VAE). The encoder and decoder that compose the generative model are trained using pure pixel information extracted directly from the observed image, what allows for an unsupervised formulation.
The proposed EM model is applied to the solution of a spectral unmixing problem, which we cast as an alternating nonlinear least-squares problem that is solved iteratively with respect to the abundances and to the low-dimensional representations of the EMs in the latent space of the deep generative model.
Simulations using both synthetic and real data indicate that the proposed strategy can outperform the competing state-of-the-art algorithms.
\end{abstract}



\section{Introduction}

Hyperspectral image analysis consists in a vast collection of algorithms and methods used to retrieve vital information from hyperspectral images (HI) in a increasing number of applications. Common applications include~\cite{kouyama2016development,Bioucas-Dias-2013-ID307} space exploration, remote sensing, surveillance, and, more recently, medical applications such as disease diagnosis and image-guided surgery~\cite{lu2014medical}.
%
One analysis methodology of particular interest is spectral unmixing (SU), which aims at retrieving sub-pixel information concerning the spectra of materials present in the scene, as well as estimating the proportions in which they contribute to each HI pixel~\cite{Keshava:2002p5667,imbiriba2018ULTRA}.

%
%
Many parametric models have been proposed to describe the interaction between light and the target surface~\cite{Keshava:2002p5667, Dobigeon-2014-ID322}.
The simplest of such models is the \textit{Linear Mixing Model} (LMM), which considers that the observed reflectance of an HI pixel is obtained from a convex combination of the spectral signatures of pure materials. This model imposes a convex geometry to the SU problem, where all HI pixels are confined to a simplex whose vertices are the pure material reflectances, usually termed endmembers (EMs). The linearity and convexity of the LMM model lead to an interpretation of its coefficients as the relative abundances of each pure material in the HI.
Nevertheless, some characteristics of practical HIs cannot be modeled by the standard LMM, such as nonlinearities~\cite{Dobigeon-2014-ID322, Imbiriba2016_tip, Imbiriba2017_bs_tip, Imbiriba2014} or variations of the EMs along the image~\cite{Thouvenin_IEEE_TSP_2016_PLMM, drumetz2016blindUnmixingELMM, imbiriba2018glmm}. More sophisticated models are required when such nonidealities have important impact on the formation of the HI.

\subsection{EM variability and learning-based SU methods}

The variation of the endmembers across an HI (also called \emph{EM variability}) is a very common effect since we can often associate multiple, different spectral signatures to each pure underlying material in a scene.
EM variability can originate from environmental conditions, illumination, or atmospheric or temporal changes~\cite{Zare-2014-ID324-variabilityReview}. Its occurrence may incur the propagation of significant estimation errors throughout the unmixing process~\cite{Thouvenin_IEEE_TSP_2016_PLMM}. Different strategies have been proposed to cope with EM variability in SU. They can be classified in methods that represent EMs as sets, methods that model EMs as statistical distributions, and methods that incorporate parametric representations of EM variability in the mixing model~\cite{drumetz2016variabilityReviewRecent}.

Parametric models are raising considerable interest since they lead to good unmixing results and avoid the main drawbacks of the other groups of SU methods that address EM variability, namely the dependence on \textit{a priori} knowledge of libraries of material spectra or the need for strong assumptions on the statistical distribution of the EMs for mathematical tractability~\cite{Zare-2014-ID324-variabilityReview,drumetz2016variabilityReviewRecent}.
Recently proposed parametric models attempt to capture spectral variability by extending the LMM using either additive~\cite{Thouvenin_IEEE_TSP_2016_PLMM} or multiplicative~\cite{drumetz2016blindUnmixingELMM,imbiriba2018glmm,Borsoi_multiscaleVar_2018,Borsoi_2018_Fusion} scaling factors, or by considering tensor-based formulations~\cite{imbiriba2018ULTRA_V,borsoi2019icassp}.
%
%

Although SU methods based on extended parametric models offer different trade-offs between representation capacity and model complexity, they still fail to achieve a desirable balance between a low-dimensional representation and enough flexibility to represent complex EM variability. Specifically, they fail to properly explore the fact that, although being very complex and spectrally non-homogeneous, spectral variability in real scenes is often confined to low-dimensional manifolds~\cite{Hapke1981,jacquemoud2001leafOpticalPropertiesReview,lobell2002moistureEffectsOnReflectance}. This property is due to the fact that the spectral signature of many materials is a function of only a few photometric or chemical properties of the medium. Prominent examples include packed particle spectra as a function of its roughness, size and density~\cite{Hapke1981}, leaf reflectance spectra as a function of various biophysical parameters~\cite{jacquemoud2001leafOpticalPropertiesReview}, and soil reflectance as a function of moisture conditions~\cite{lobell2002moistureEffectsOnReflectance}. Thus, existing models tend to be either too restrictive in their modeling capability or to lead to severely ill-posed estimation problems.

SU considering EM variability has also been formulated as a supervised learning problem, which is then solved without the need for an accurate physical model using neural networks (NNs) or support vector machines (SVMs)~\cite{okujeni2013learningSVRunmixingUrban,wang2013unmixingSVMregressionResidueConstr,mianji2011unmixingVariabilitySVMclassification,parente2017unmixingGenModels}.
%
However, these strategies depend on the availability of vast amounts of training data to adequately capture the spectral diversity of real scenes. This makes the training process computationally intensive and often intractable for large EM libraries, which must also be known \emph{a priori}.
Some works attempt to reduce the computational cost of these solutions by modifying learning algorithms to use hybrid soft-hard classification~\cite{wang2009unmixingExtendedSVMpurePixels,gu2013unmixingExtendedSVMpurePixelsMultiKernel,li2015unmixingExtendedSVMgeometricAnalysis}. However, the resulting reconstructed abundance fractions do not have a clear physical interpretation due to the lack of a direct relationship to a physically motivated mixing model. 

More recently, unsupervised SU approaches have also emerged by using autoencoders (AEC), which consist of encoder-decoder structured  NNs originally devised for nonlinear dimensionality reduction~\cite{van2009dimensionalityReductionRev}. 
%
These methods attempt to associate the decoder structure of the network with the LMM and the low-dimension representation of the input spectral vectors to the fractional abundances~\cite{palsson2018autoencoderUnmixing_IEEEaccess}. 
Different variations have been proposed, using pre-processing steps to reduce noise and outliers~\cite{guo2015autoencodersUnmixing,su2018autoencodersUnmixing}, untiying the decoder from the encoder weights~\cite{qu2018udas_autoencoderUnmixing}, using spectral angle distances to address nonlinear SU~\cite{ozkan2018endnet_autoencoderUnmixing},  or using denoising autoencoders to generate a robust initialization to matrix factorization-based SU strategies~\cite{su2019deepAutoencoderUnmixing}.
In~\cite{ozkan2018autoencoderUnmixingVariability} the authors proposed a nonlinear encoder-decoder structure to address the unmixing problem considering spectral variability. The proposed solution involves the simultaneous training of six neural networks to optimize a very large number of parameters. An autoencoder structure is employed to estimate the parameters of a spectral model of a hyperspectral image by minimizing the image reconstruction error while limiting the energy of some of the model parameters. No regularization strategy connecting the different pixels of the image is employed.


%
Despite their popularity, supervised learning-based SU algorithms are still not able to properly address the spectral variability problem, as they depend on extremely large amounts of labeled training data, leading to a computationally unfeasible learning process. Furthermore, the lack of a clear connection between AEC-based strategies and the physical mixing process makes one skeptic when concerning the robustness of AEC-based SU in face of more complex phenomena such as spectral variability.

\subsection{Proposed methodology}

In this work, we propose a novel SU formulation that leverages the advantages of deep learning methods to address EM variability while still maintaining a strong connection to the physical mixing process, and using limited amounts of training data.
Specifically, we adopt a deep generative NN to represent the manifold of EM spectra, which is then incorporated within the LMM. 
%
%
Generative models such as variational autoencoders (VAE) \cite{kingma2013AEC_varBayes} and generative adversarial networks (GAN) \cite{goodfellow2014GANs} have recently obtained excellent performance at learning the probability distribution of complex data sets in very high dimensional spaces (e.g. natural images) from relatively small amounts of training data. The structure of generative models allow one to find a low-dimensional latent representation that parsimonously describes the variability of complex high-dimensional data sets. This leads to a low-dimensional parametrization of the training data distribution.

We formulate a novel unmixing strategy that can be cast as the problem of estimating the latent representations of the generative endmember models and the corresponding fractional abundances for each pixel in the HI.
Specifically, we break down the SU problem in two steps.
In the first step, we learn the latent EM variability manifold for each material in the scene using a deep generative EM model. The learning process uses pure pixel information directly extracted from the observed HI, which makes the proposed strategy suitable for unsupervised SU. 
In the second step, an alternating least-squares strategy is employed to estimate the parameters of an extended version of the LMM parametrized using the generative EM models obtained in the first step. The corresponding optimization problem is solved iteratively with respect to the abundances and to the low-dimensional representations of the EMs in the latent space of the deep generative models.

%
As a result, the proposed approach benefits from the reduced dimension of the latent space. Moreover, unlike current approaches, the new method does not depend on the careful selection of regularization parameters to yield a good performance. The resulting algorithm is named \emph{Deep Generative Unmixing algorithm} (DeepGUn).
The proposed method is strongly related to parametric models and leverages the learning and generalization capability of deep neural networks to properly represent the manifold of EM variability. Hence, DeepGUn leads to a model that is both low-dimensional and physically accurate, better describing the variability actually present in the scene.



%

Experimental results performed with both synthetic and real data indicate that the proposed strategy leads to more accurate abundance estimations than standard state-of-art SU methods accounting for EM variability. Qualitative analysis of the estimated abundance maps confirms these results. The improved accuracy comes at the expense of a small increase in the computational cost when compared to the best competing strategies.

This paper is organized as follows. Section~\ref{sec:lmms} briefly reviews the LMM and its parametric extended versions. Section~\ref{sec:gen_models} discusses the basic properties of generative models in the context of VAE and GAN. Section~\ref{sec:prop} introduces the proposed generative EM model and its learning strategy. In Section~\ref{sec:Problem} we formulate the resulting SU problem, present the DeepGUn algorithm, and discuss aspects of the proposed optimization strategy. The neural network architecture is discussed in Section~\ref{sec:NNA}. The performance of the proposed method is compared with that of competing algorithms in Section~\ref{sec:Simulations}. Finally, the conclusions are presented in Section~\ref{sec:conclusions}.

\section{Linear Mixing Models} \label{sec:lmms}
The Linear Mixing Model (LMM)~\cite{Keshava:2002p5667} assumes that a given $n$-th pixel $\by_n\in\amsmathbb{R}^L$, with $L$ bands, is represented as
\begin{align}
 &\by_n = \bM \ba_n + \be_n, \label{eq:LMM}
\quad \text{subject to }\,\cb{1}^\top\ba_n = 1 \text{ and } \ba_n \geq \cb{0} 
\end{align}
where $\bM \in \amsmathbb{R}^{L\times P}$ is a matrix whose columns are the $P$ EM spectral signatures $\bm_k$, $\ba_n$ is the abundance vector and $\be_n$ is an additive white Gaussian noise (WGN).
The LMM assumes that the EM spectra are fixed for all HI pixels $\by_n$, $n=1,\ldots,N$. 
This assumption jeopardizes the accuracy of estimated abundances in many circumstances due to the spectral variability existing in a typical scene.

Different parametric models have been recently proposed to account for variable EM spectra within a given scene~\cite{drumetz2016blindUnmixingELMM,Thouvenin_IEEE_TSP_2016_PLMM, imbiriba2018glmm}. These models can be generically described as
\begin{align}
\label{eq_generic_model}
    \by_n {}={} f(\bM_0,\btheta_n)\ba_n+\be_n
\end{align}
where $f$ is a parametric function, $\bM_0\in\amsmathbb{R}^{L\times P}$ is a reference EM matrix, and $\btheta_n$ is a vector of parameters describing the manifold of EM variability.

Different functional forms have been proposed for $f(\bM_0,\btheta_n)$ to account for EM variability in this framework, such as additive~\cite{Thouvenin_IEEE_TSP_2016_PLMM} or multiplicative~\cite{drumetz2016blindUnmixingELMM, imbiriba2018glmm} variability factors acting upon the reference EM matrix~$\bM_0$.
However, these models fail to achieve a desirable balance between a low-dimensional representation and enough flexibility to represent complex variability patterns. They tend to be either too restrictive in their modeling capability, or to lead to ill-posed optimization problems~\cite{borsoi2019icassp}. Instead of using a pre-defined parametric model, we propose to address this issue by learning a parametric function $f(\bM_0,\btheta_n)$ using a generative model.

\section{Generative Models} \label{sec:gen_models}

Generative models attempt to estimate the probability distribution $p(X)$ of a random variable $X\in\amsmathbb{R}^L$ based on a set of observations $\bx_i$, $i=1,\ldots,N_x$ in such a way that allows one to generate new samples that look similar to new realizations of~$X$.
The main characteristic of this problem, which sets it apart of other unsupervised learning methods such as density estimation, is the fact that we must to be able to sample from the estimated model $\hat{p}(X)$.

In many practical applications of interest, the dimensionality~$L$ of the variable of interest $X$ is very high. This makes the general problem very difficult, as it amounts to estimating and sampling from an arbitrary high-dimensional probability density function~\cite{neal2001annealedDensityEstimation,arjovsky2017wassersteinGANs}.
Nonetheless, the distributions of interest are often supported at a low-dimensional manifold of a set of so-called latent variables, and this fact can be explored to make the problem more tractable.
A convenient way to address this problem is to define a new random variable $\amsmathbb{R}^K\ni Z\sim p(Z)$ with a known distribution in a low-dimensional space (e.g. an isotropic Gaussian distribution with $K\ll L$), and a parametric function (e.g. a neural network) $\mathcal{G}_{\theta}$ mapping $Z\mapsto \widehat{X}\in\amsmathbb{R}^L$ such that the image of $Z$ by $\mathcal{G}_{\theta}$ is a random variable whose distribution is very close to $p(X)$. 
In other words, the goal becomes to learn the parameters $\theta$ of $\mathcal{G}_{\theta}$ such that the distribution of $\widehat{X}=\mathcal{G}_{\theta}(Z)$ is as close to $p(X)$ as possible.
Then, samples of $\widehat{X}$ can be generated by sampling from $Z\sim p(Z)$ and using the mapping $\mathcal{G}_{\theta}(Z)$.

%
Although estimating $\theta$ may still seem difficult at first, recent advances in machine learning such as VAEs~\cite{kingma2013AEC_varBayes} and GANs~\cite{goodfellow2014GANs} have shown formidable performance at learning complex distributions such as those of natural images.

VAEs address this problem by assuming that the distribution of the observed data $X$ follows a directed graphical model $p(X|Z)$, which is represented by the function~$\mathcal{G}_\theta$. The parameters of $\mathcal{G}_\theta$ are learned by maximizing a lower bound on the log-likelihood of $p(X)$~\cite{kingma2013AEC_varBayes}:
\begin{align} \label{eq:VAE_cf}
    \log\,p(X) {}\geq{} &
    \Ex_{q_\phi(Z|X)}\big\{ \log p(X|Z) \big\}
    \nonumber \\ &
    - KL\big(q_\phi(Z|X)\|p(Z)\big)
\end{align}
where $KL(\cdot\|\cdot)$ is the Kullback-Leibler divergence between two distributions, $\Ex_{\varsigma}\{\cdot\}$ is the expected value operator with respect to the distribution  $\varsigma$ and $q_\phi(Z|X)$ is a variational approximation to the intractable posterior $p(Z|X)$, which is modeled by a function $\mathcal{D}_\phi$ (e.g. another  neural network)  parameterized in $\phi$. Note that $q_\phi(Z|X)$ must be a high-capacity distribution\footnote{Capacity of a distribution is a generic term to describe how complex a relationship it can model.}, so that it can provide a good approximation of the posterior $p(Z|X)$, which then allows the lower bound in~\eqref{eq:VAE_cf} to be close to the true value of $\log\,p(X)$~\cite{doersch2016tutorialVAE}.

GANs, on the other hand, attempt to learn the distribution $p(X)$ by searching for the Nash equilibrium of a two-player adversarial game~\cite{goodfellow2014GANs}. A generator network $\mathcal{G}_\theta$ tries map the distribution of the latent variables $Z$ into the data distribution of $X$, and a discriminator network $\mathcal{C}_\phi$ tries to predict the probability of a random sample $\bx_i$ coming from the true distribution $p(X)$ instead of being generated through~$\mathcal{G}_\theta$.
The generator $\mathcal{G}_\theta$ is trained by maximizing the probability of the discriminator making a mistake. This is formulated as the minimax optimization problem
\begin{align} \label{eq:GANs_opt_cf}
    \min_{\mathcal{G}_\theta} \, \max_{\mathcal{C}_\phi} \,\,\,  &
    \Ex_{p(X)}\big\{\log \mathcal{C}_{\phi}(X) \big\}
    \nonumber \\ &
    + \Ex_{p(Z)}\big\{(1-\mathcal{C}_{\phi}(\mathcal{G}_\theta(Z)))\big\}
    \,.
\end{align}
GANs are more flexible and have shown better performance at approximating complex distributions such as natural images (leading to sharper results) when compared to VAEs~\cite{goodfellow2014GANs}. However, GANs are also much harder to train~\cite{arjovsky2017wassersteinGANs}. Moreover, VAEs naturally offer a way to obtain the latent representations corresponding to samples $\bx_i\sim p(X)$ by mapping $X\mapsto Z$ using the function $\calD_\phi$, which is also called an encoder model.
This property and their stable training have motivated us to use VAEs in this work.






\begin{figure*}
    \centering
    \includegraphics[width=0.9\linewidth]{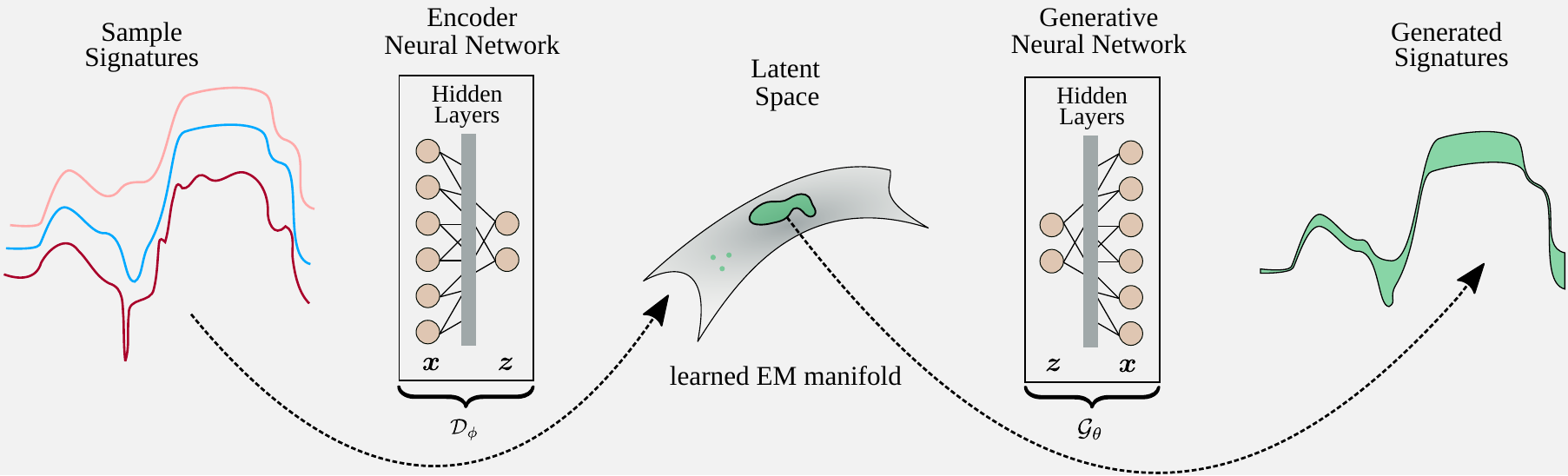}
    \caption{Illustration of the proposed Deep Generative Endmember Model.}
    \label{fig:GEM}
\end{figure*}

\section{A Deep Generative Endmember Model} \label{sec:prop}

In this section, we propose to model the distribution of EM spectral variability using a deep generative model.
By doing so, we can explicitly explore a common property of spectral variability: the EM spectra are usually confined to a low-dimensional manifold. This property is due to the fact that the spectral signature of many materials is a function of a few photometric or chemical properties of the medium. Prominent examples include packed particle spectra as a function of its roughness, size and density~\cite{Hapke1981}, leaf reflectance spectra as a function of various biophysical parameters~\cite{jacquemoud2001leafOpticalPropertiesReview}, and soil reflectance as a function of moisture conditions~\cite{lobell2002moistureEffectsOnReflectance}.

\subsection{The steps of the proposed SU method}

We assume the existence of nonlinear functions $\calG_{\theta_p}$, $p=1,\ldots,P$ (the generative model) that map latent representations $\bz_p$ into their corresponding spectral signatures $\bm_p$. We assume also the existence of encoder models $\calD_{\phi_p}$ that map spectral signatures into their latent representations.
In other words, we assume that any arbitrary observation $\bm_p$ of a spectral signature of a material belongs to the set
\begin{align}
    \bm_p \in \big\{\calG_{\theta_p}(\bz_p)\,:\,\bz_p\in\amsmathbb{R}^K\big\}
\end{align}
and thus can be equivalently represented by a corresponding low-dimensional vector $\bz_p\in\amsmathbb{R}^{K}$ in the latent space of the generative model~$\calG_{\theta_p}$. 
This reasoning is illustrated in Fig.~\ref{fig:GEM}, where the encoder function $\calD_\phi$ maps the input EM signature to the low-dimensional manifold. Reciprocally, low-dimensional vectors in the latent space can be mapped (decoded) to their corresponding spectral signatures using $\calG_\theta$.

As such, we can formulate EM estimation in the SU problem in the latent domain (as opposed to the input spectral space), which is of a much lower order. Moreover, this approach will keep the physical  interpretation of the model, provided that we have relevant training data to learn the generative models~\cite{shah2018invGANpriors,bora2017compressedInvGANs,anirudh2018unsupervisedInverseProblemsGANs,asim2018GANsDeblurring}.
This strategy relies on the existence of a priori training data for each material in the image, which might come in the form of, e.g., spectral libraries of laboratory measurements~\cite{iordache2011sunsal}. Nevertheless, we propose a more practical and effective approach to train the generators~$\calG_{\theta_p}$ by exploring information contained in multiple pure pixels extracted from the observed HI. The presence of multiple pure pixels in an observed HI is a characteristic of many real scenes, and can be leveraged to help in estimating the EM models, thus, reducing the ill-posedness of the SU problem\footnote{Pure pixels are defined here as a set of pixels whose spectral distance relative to the reference EMs in $\bM_0$ is less than a specified threshold.}.

Therefore, we propose to break the unmixing problem into a sequence of two problems: 
\begin{itemize}
    \item[i)] Using pure pixel information extracted from the HI by standard EM extraction methods, learn the generative and encoder models, $\calG_{\theta_p}$ and $\calD_{\phi_p}$, for all EMs in the scene ($p=1,\ldots,P$).
    
    \item[ii)] Using the learned generative models, solve the SU problem by estimating the latent EM representations $\bZ_n=[\bz_{1,n},\ldots,\bz_{P,n}]$ and the fractional abundance vectors $\ba_n$ that can best represent the observed hyperspectral data, for all pixels in the scene ($n=1,\ldots,N$).
\end{itemize}


\subsection{Learning the generative and encoder models~$\calG_{\theta_p}$ and~$\calD_{\phi_p}$}

The objective of this first problem is to estimate the generative and encoding models $\calG_{\theta_p}$ and $\calD_{\phi_p}$, for $p=1,\ldots,P$. We assume the knowledge of a set $\mathcal{N}_{\mathcal{P},p}$ of pure pixels for the $p$-th EM, for all $p=1,\ldots,P$. Multiple pure pixels exist in many scenes, and can be directly extracted from the observed HI using automated EM extraction techniques~\cite{somers2012automatedBundlesRansomSampling,somers2016endmemberLibrariesChapter}.
The sets of pure pixels $\mathcal{N}_{\mathcal{P},p}$, which can be seen as observations from the statistical distribution of each EM, are then used in the form of training data to learn the models $\calG_{\theta_p}$ and $\calD_{\phi_p}$ using a VAE~\cite{kingma2013AEC_varBayes}.
If the set $\mathcal{N}_{\mathcal{P},p}$ is representative of the variability of the $p$-th material, the learned generative model $\calG_{\theta_p}$ will be able to accurately describe the manifold of the $p$-th EM variability. Doing the same for all $p=1,\dots,P$ yields a set of variability models for all the EM spectra.

Although the extraction of multiple pure pixels from observed HIs is a well-established technique used to produce EM libraries~\cite{somers2012automatedBundlesRansomSampling}, mixed pixels can sometimes be mistakenly identified as a pure pixel of some of the EMs. This constitutes a problem for library-based SU applications (e.g. MESMA and sparse SU) since some of the library spectra may end up not being representative of their EM class (material).

The smooth nature of the latent representation of VAEs allows the mitigation of this problem in the proposed approach.
%
%
Assuming the availability of a reference EM matrix $\bM_0$ of correctly identified signatures (which can be obtained using any EM extraction method) and of a set of encoder models $\calD_{\phi_p}$, we can compute the latent representation of these reference signatures of each EM as 
\begin{align} \label{eq:Z0_def}
    \bZ_0  &  {}={} \big[\bz_{1,0},\ldots,\bz_{P,0}\big]
    \nonumber \\ & 
    {}={} \big[\calD_{\phi_1}(\bm_{1,0}),\ldots,\calD_{\phi_P}(\bm_{P,0})\big]
    \,.
\end{align}
where $\bm_{p,0}$ is the $p$-th column of $\bM_0$. The latent representation $\bz_{p,0}$ can be used as a reference latent code for the $p$-th material.
%
Thus, we can measure how close an estimated EM latent representation~$\bz_p$ is to the latent representation of a pure pixel by evaluating its Euclidean distance to~$\bz_{p,0}$.
This can be performed since the output of VAEs have been shown to vary smoothly with changes of the latent variable~\cite{kingma2013AEC_varBayes}.
Thus, we can use $\bZ_0$ to regularize the SU problem to prevent $\calG_{\theta_p}(\bz_p)$ from representing mixed pixels. This increases the robustness of the proposed approach.

\subsection{Extracting sets of pure pixels from the observed HI}

An important part of the proposed methodology consists in the extraction of the sets of pure pixels~$\mathcal{N}_{\mathcal{P},p}$, $p=1,\ldots,P$ from the observed hyperspectral image~$\bY$. Although different strategies have been proposed for image-based library construction (see e.g.~\cite{uezato2016novelBundlesExtractionClustering,andreou2016bundlesExtractionVariabilityMultiscaleBand}), these techniques depend on multiple parameters that must be carefully adjusted in order to obtain good results.
%
Instead of these approaches, we adopt a very simple strategy to select pure pixels from an HI that makes use of the reference matrix~$\bM_0$ extracted from the image using a pure-pixel-based endmember extraction algorithm (e.g.~VCA~\cite{Nascimento2005}), which will also later be used to construct $\bZ_0$ in~\eqref{eq:Z0_def}. We simply select as the elements of~$\mathcal{N}_{\mathcal{P},p}$ the $S_p$ image pixels that have the smallest spectral angle to the reference signature in the $p$-th column of~$\bM_0$, where $S_p$ is the cardinality of~$\mathcal{N}_{\mathcal{P},p}$ for $p=1,\ldots,P$.
Although the success of this strategy depends on having a reasonably accurate estimation of~$\bM_0$, we experimentally found it to be more robust and easier to adjust than, for instance,  the one in~\cite{somers2012automatedBundlesRansomSampling}.

\section{The Unmixing Algorithm} \label{sec:Problem}
Given a set of generative models $\calG_{\theta_p}:\amsmathbb{R}^{K}\to\amsmathbb{R}^L$, $p=1,\ldots,P$ for each EM in the scene, a latent space representation $\bZ_0$ of a reference EM matrix $\bM_0$, and an HI $\bY = [\by_1,\ldots,\by_N]$, the SU problem can be cast as the minimization of a risk functional of the form
\begin{equation} \label{eq:general_unmxing_cf}
    J(\bA,\bbZ) {}={}  \frac{1}{2} \sum_{n=1}^N \|\by_n - \widetilde{\calG}(\bZ_n) \ba_n\|_F^2 
    + \mathcal{R}(\bA) + \mathcal{R}(\bbZ)
\end{equation}
where $\bA = [\ba_1, \dots, \ba_N]\in\amsmathbb{R}^{P\times N}$ is the abundance matrix, $\bbZ\in\amsmathbb{R}^{N\times P\times K}$ is a 3-D tensor obtained by stacking all pixel-dependent latent EM representations $\bZ_n$, such that $[\bbZ]_{n,:,:}=\bZ_n$, $\mathcal{R}(\bA)$ and $\mathcal{R}(\bbZ)$ are regularization terms to improve the problem conditioning, and the matrix-valued function $\widetilde{\calG}(\bZ_n)$ defined as
\[\widetilde{\calG}(\bZ_n) = \big[\calG_{\theta_1}(\bz_{1,n}),\ldots,\calG_{\theta_P}(\bz_{P,n})\big], \,\,\, n=1,\ldots,N\]
is the concatenation of the generative functions for each EM.

The term $\mathcal{R}(\bA)$ is a regularization functional that aims to provide spatial smoothness and to enforce positivity and sum-to-one constraints to the abundances. It is given by~\cite{imbiriba2018glmm}
\begin{align}
	\mathcal{R}(\bA) {}={} & \lambda_A \big(\|\mathcal{H}_h(\bA)\|_{2,1} + \|\mathcal{H}_v(\bA)\|_{2,1} \big)
    + \iota_{\mathcal{S}^1}(\bA)
\end{align}
where parameter $\lambda_A$ controls the contribution of this term to the cost function.
The first two terms are a spatial regularizers over $\bA$, where $\mathcal{H}_h$ and $\mathcal{H}_v$ are linear operators that compute the first-order horizontal and vertical gradients of a bidimensional signal, acting separately for each material of $\bA$, and $\|\cdot\|_{2,1}$ is the $\mathcal{L}_{2,1}$ norm, defined as $\|\bX\|_{2,1}=\sum_{n=1}^N\|\bx_n\|_2$. 
The term $\iota_{\mathcal{S}^1}(\bA)$ is the indicator function of the unity simplex, i.e. $\iota_{\mathcal{S}^1}(\bA)=0$ if $\bA\in\mathcal{S}^1$ and $\iota_{\mathcal{S}^1}(\bA)=\infty$ otherwise, where
\begin{align}
    \mathcal{S}^1 {}={} \big\{ \bA\in\amsmathbb{R}^{P\times N} \,:\, \bA\geq0,\, \cb{1}^\top\bA = \cb{1}^\top \big\} 
    \,.
\end{align}
The term $\mathcal{R}(\bbZ)$ constrains the EM latent representations $\bbZ$ to be close to the latent representation $\bZ_0$ of the reference EM matrix $\bM_0$. It is given by
\begin{align}
    \mathcal{R}(\bbZ) {}={} \frac{\lambda_Z}{2} \sum_{n=1}^N \|\bZ_n-\bZ_{0}\|_F^2
\end{align}
where parameter $\lambda_Z$ controls the contribution of this term to the cost function.
This regularization makes the estimation problem more robust to the selection of the training data~$\mathcal{N}_{\mathcal{P},p}$ by assuring the closeness of the estimated latent codes $\bbZ$ and the representations of pure pixels of each class.
However, it relies indirectly on the reference EM signatures~$\bM_0$ (which are extracted from the observed HI with endmember extraction algorithms) being adequate representatives of their material classes in order to provide a good performance.

The optimization problem then becomes
\begin{equation} \label{eq:opt_deepGen}
    (\,\widehat{\!\bA},\widehat{\bbZ}) {}={}
    \mathop{\arg\min}_{\bA,\,\cb{\bbZ}} J(\bA,\cb{\bbZ}).
\end{equation}
The problem defined in~\eqref{eq:opt_deepGen} is non-smooth and non-convex if solved simultaneously with respect to both variables $\bA$, and $\bbZ$. However, an approximate solution can be found by minimizing~\eqref{eq:opt_deepGen} iteratively with respect to each variable, leading to the Deep Generative Unmixing (DeepGUn) method described in Algorithm~\ref{alg:global_opt}.
The DeepGUn algorithm consists of two distinctive steps. First, the generative endmember models generative and encoder models $\calG_{\theta_p}$, $\calD_{\phi_p}$, $p=1,\ldots,P$ are trained based on the pure pixels $\mathcal{N}_{\mathcal{P},p}$, $p=1,\ldots,P$ extracted from the observed HI and $\bZ_0$ is computed. Afterwards, the alternating minimization approach is applied to compute the abundance maps and the latent representations of the endmembers for each pixel.
We next describe the details of each optimization step. Implementation details are described in Sections~\ref{sec:NNA} and \ref{sec:Simulations}.

\begin{algorithm} [bth]
\SetKwInOut{Input}{Input}
\SetKwInOut{Output}{Output}
\caption{DeepGUn algorithm for solving~\eqref{eq:opt_deepGen}\label{alg:global_opt}}
\Input{$\bY$, $\lambda_Z$, and $\lambda_A$.}
\Output{$\,\widehat{\!\bA}$ and $\widehat{\bbM}$.}
Estimate the reference EM signatures $\bM_0$ using an EM extraction method (e.g.~VCA)\;
Estimate $\bA^{(0)}$ using a standard LMM-based SU method\;
Extract sets of pure pixels $\mathcal{N}_{\mathcal{P},p}$, $p=1,\ldots,P$ from the HI using a bundle extraction strategy\;
Train the generative and encoder models $\calG_{\theta_p}$, $\calD_{\phi_p}$, $p=1,\ldots,P$ \;
Compute the latent representation of $\bM_0$ as $\bZ_0=[\calD_{\phi_1}(\bm_{1,0}),\ldots,\calD_{\phi_P}(\bm_{P,0})]$ \;
Set $i=0$ \;
\While{stopping criterion is not satisfied}{
$i=i+1$ \;
$\bbZ^{(i)} = \underset{\bbZ}{\arg\min} \,\,\,\,  {J}(\bA^{(i-1)},\bbZ)$ \;
$\bA^{(i)} = \underset{\bA}{\arg\min} \,\,\,\,  {J}(\bA,\bbZ^{(i)})$ \;
}
\textbf{for} $n=1,\ldots,N$, \textbf{do}, $[\widehat{\bbM}]_{:,:,n}=\widetilde{\calG}([\bbZ]_{:,:,n})$, \textbf{end}\;
\KwRet $\,\widehat{\!\bA}=\bA^{(i)}$,~ $\widehat{\bbM}$ \;
\end{algorithm}

\subsection{Optimization with respect to $\bbZ$}
Rewriting~\eqref{eq:opt_deepGen} considering only the terms in~\eqref{eq:general_unmxing_cf} that depend on $\bbZ$, the problem becomes
\begin{equation} \label{eq:Z_problem}
\begin{split}
    \min_{\bbZ} \,\,\, \frac{1}{2} \sum_{n=1}^N \Big(\|\by_n - \widetilde{\calG}(\bZ_n) \ba_n\|_F^2 
    + \lambda_Z \|\bZ_n-\bZ_{0}\|_F^2 \Big)
\end{split}
\end{equation}
This is a regularized nonlinear least squares problem, which can be solved individually for each pixel~$\by_n$.
%
%
Thus,~\eqref{eq:Z_problem} can be decomposed into $N$ non-convex, nonlinear optimization problems with dimensionality $K\times P$ by denoting each summand in~\eqref{eq:Z_problem} by $\widetilde{J}^{\,(n)}$, $n=1,\ldots,N$. We solve each of those problems~$\widetilde{J}^{\,(n)}$ using a quasi-Newton algorithm, described in Algorithm~\ref{alg:latent_opt}, which provides an efficient solution for high-dimensional functions $\widetilde{\calG}$~\cite{nocedal2006optimizationBook}.

Although problem~\eqref{eq:Z_problem} is generally non-convex, recent research~\cite{hand2018convexityInverseOptGANs} has proven that, under suitable assumptions on the generator network $\widetilde{\calG}$, the problem of recovering the latent variable $\bZ_n$ does not have any stationary point (e.g. local minima or saddle points) outside a small neighborhood of the desired solution and its negative scalar multiple. This indicates the existence of a favorable global geometry of~\eqref{eq:Z_problem}.


Note that $\widetilde{\calG}$ is not necessarily differentiable with respect to the latent representations~$\bZ_n$, which can make the optimization problem more challenging. Nonetheless, quasi-Newton algorithms show excellent performance at non-smooth problems~\cite{lewis2013nonsmoothQuasiNewton}, where convergence is generally observed as long as the line search procedure does not return a point at which the objective function is non-differentiable. This allows quasi-Newton algorithms to be directly applied to obtain approximate solutions to non-smooth problems with  good computational efficiency~\cite{lewis2013nonsmoothQuasiNewton,curtis2015quasiNewtonNonsmooth}.

\begin{algorithm} [bth]
\SetKwInOut{Input}{Input}
\SetKwInOut{Output}{Output}
\caption{Quasi-Newton algorithm for solving~\eqref{eq:Z_problem}~\label{alg:latent_opt}}
\Input{$\ba_n$, $\by_n$, $\lambda_Z$, $\bZ_0$ and $\widetilde{J}^{\,\,(n)}$.}
\Output{$\bZ_n$.}
Set $i=0$ and $\bB_1=\bI$ \;
\While{stopping criterion is not satisfied}{
$i=i+1$ \;
Compute search direction $\bp_i=-\bB_i\nabla \widetilde{J}_{i}^{\,\,(n)}$ \;
Set $\bz_{i+1}=\mu_i\bp_i$, where $\mu_i$ is computed using a line search procedure to satisfy the Wolfe conditions\;
Define $\bs_i=\bz_{i+1}-\bz_{i}$ and $\bu_i=\nabla \widetilde{J}_{i+1}^{\,\,(n)}-\nabla \widetilde{J}_{i}^{\,\,(n)}$\;
$\bB_{i+1} = \bB_i - \frac{\bB_i\bs_i\bs_i^\top\bB_i}{\bs_i^\top\bB_i\bs_i} + \frac{\bu_i\bu_i^\top}{\bu_i^\top\bs_i}$ \;
}
Reorder $\bz_{i+1}$ as a matrix $\bZ_n$  \;
\KwRet $\widehat{\bZ}_n=\bZ_n$ \;
\end{algorithm}

\subsection{Optimization with respect to $\bA$}
Restating~\eqref{eq:opt_deepGen} considering only the terms in~\eqref{eq:general_unmxing_cf} that depend on $\bA$ leads to
\begin{equation} \label{eq:A_problem}
\begin{split}
    \min_{\bA} \,\,\, &
    \frac{1}{2} \sum_{n=1}^N \|\by_n - \widetilde{\calG}(\bZ_n) \ba_n\|_F^2 
    + \iota_{\mathcal{S}^1}(\bA)
    \\ &
    + \lambda_A ( \|\cp{H}_h(\bA)\|_{2,1} + \|\cp{H}_v(\bA)\|_{2,1})
    \,.
\end{split}
\end{equation}
Since the latent variables $\bZ_n$ are fixed,~\eqref{eq:A_problem} consists of a SU problem with a pixel-dependent EM matrix and an edge-preserving spatial regularization. Although this problem is not separable with respect to each pixel in the image, the Alternating Direction Method of the Multipliers (ADMM) can be used to obtain an efficient solution~\cite{Boyd_admm_2011}. The solution of~\eqref{eq:A_problem} using the ADMM is well described elsewhere (e.g. \cite{drumetz2016blindUnmixingELMM}) and will thus be omitted here for conciseness.

\section{Neural Network Architecture}~\label{sec:NNA}

As discussed before, we used a VAE~\cite{kingma2013AEC_varBayes} to learn the generative and encoder models $\calG_{\theta_p}$ and $\calD_{\phi_p}$ from the sets of pure pixels $\mathcal{N}_{\mathcal{P},p}$. Compared to GANs, the training of VAEs is much simpler and more stable~\cite{arjovsky2017wassersteinGANs}. Moreover, VAEs naturally return the encoder model $\calD_{\phi_p}$ as an approximation to the posterior distribution when learning $\calG_{\theta_p}$.
We have selected a dimension $K=2$ for the latent space, as it was experimentally verified to be sufficient to adequately capture the variability of each single material in a scene.

For the network architectures, we selected the number of layers and neurons according to the autoencoder implementation in~\cite{van2008matlabToolbox,van2009dimensionalityReductionRev}, with three hidden layers using ReLU activations (defined as $ReLU(x)=\max(x,0)$) in the hidden layers, which are described in more detail in Tables~\ref{tab:encoder_architecture} and~\ref{tab:decoder_architecture}.

We found that this configuration led to spectrally smooth generated signatures, and was effective at generalizing well with small training sample sizes.
%
We trained the network for 50 epochs with the Adam optimizer~\cite{kingma2014adam} in TensorFlow, using a batch optimization with mini-batch size equal to one third of the total amount of training data for each EM.

\begin{table}
\small
\centering
\renewcommand{\arraystretch}{1.2}
\caption{Encoder network architecture.}
\vspace{-0.2cm}
\begin{tabular}{ccc}
    \hline 
    Layer       & Activation Function   & Number of units \\ \hline 
    Input       &  ---  & $L$ \\
    Hidden \# 1 & ReLU  & $\lceil 1.2\times L \rceil + 5$ \\
    Hidden \# 2 & ReLU  & $\max\big\{\lceil L/4 \rceil,\, K+2\big\} + 3$ \\
    Hidden \# 3 & ReLU  & $\max\big\{\lceil L/10 \rceil,\, K+1\big\}$ \\
    \hline 
\end{tabular}
\label{tab:encoder_architecture}
\end{table}
\begin{table}[h]
\small
\centering
\renewcommand{\arraystretch}{1.2}
\caption{Decoder network architecture.}
\vspace{-0.2cm}
\begin{tabular}{ccc}
    \hline 
    Layer        & Activation Function   & Number of units \\ \hline 
    Hidden \# 1  & ReLU & $\max\big\{\lceil L/10 \rceil,\, K+1\big\}$ \\
    Hidden \# 2  & ReLU & $\max\big\{\lceil L/4 \rceil,\, K+2\big\} + 3$ \\
    Hidden \# 3  & ReLU & $\lceil 1.2\times L \rceil + 5$  \\
    Output       & Sigmoid & $L$ \\
    \hline 
\end{tabular}
\label{tab:decoder_architecture}
\end{table}

\section{Experimental Results} \label{sec:Simulations}

In this section, simulation results using both synthetic and real data illustrate the performance of the proposed method. We compare the proposed DeepGUn method with the fully constrained least squares (FCLS), the PLMM~\cite{Thouvenin_IEEE_TSP_2016_PLMM}, the ELMM~\cite{drumetz2016blindUnmixingELMM}, and the GLMM~\cite{imbiriba2018glmm}.
In all experiments, the VCA algorithm~\cite{Nascimento2005} was used to extract the reference EM matrix $\bM_0$ from the observed HI and to initialize the different methods. The abundance maps of all methods were initialized using the results obtained by the~FCLS algorithm. The sets $\mathcal{N}_{\mathcal{P},p}$ of pure pixels were constructed by selecting the~$100$ image pixels $\by_n$ with the smallest spectral angles relative to the reference EMs in $\bM_0$. We ran the alternating optimization process in Algorithm~\ref{alg:global_opt} for at most~$10$ iterations or until the relative change of~$\bA$ and~$\bbZ$ was less than~$10^{-3}$. The iterative procedure in Algorithm~\ref{alg:latent_opt} was run until the relative change of $\bz_{i}$ was less than~$10^{-3}$. The performances were evaluated using the Normalized Root Means Squared Error (NRMSE) between the estimated abundance maps ($\text{NRMSE}_{\bA}$), between the EM matrices ($\text{NRMSE}_{\bbM}$) and between the reconstructed images ($\text{NRMSE}_{\bY}$).
The NRMSE between a true, generic tensor~$\bbX$ and its estimate~$\widehat{\bbX}$ is defined as
\begin{equation}
    \text{NRMSE}_{\bbX} = \sqrt{\frac{\|\bbX- \widehat{\bbX}\|^2_F}{\|\bbX\|^2_F}} 
    \,.
\end{equation}
Note that for the case of $\text{NRMSE}_{\bY}$, the reconstructed image $\widehat{\bY}$ is given by $[\widehat{\bY}]_{:,n}=[\widehat{\bbM}]_{:,:,n}[\,\widehat{\!\bA}]_{:,n}$, $n=1,\ldots,N$.

We consider also the Spectral Angle Mapper (SAM) to evaluate the estimated EMs
\begin{equation}
  \text{SAM}_{\bbM} = \frac{1}{N}\sum_{n=1}^{N}\sum_{p=1}^{P}\arccos\bigg(\frac{\bm_{p,n}^\top\widehat{\bm}_{p,n}}{\|\bm_{p,n}\|\|\widehat{\bm}_{p,n}\|}\bigg).
\end{equation}
where $\bm_{p,n}$ and $\widehat{\bm}_{p,n}$ are the true and the estimated signatures of the $p$-th endmember in the $n$-th pixel, respectively.

\begin{table}
\small
\caption{Simulation results using synthetic data.}
\vspace{-0.2cm}
\centering
\renewcommand{\arraystretch}{1.2}
\setlength{\tabcolsep}{3.3pt}
\begin{tabular}{lcccccc}
\bottomrule
\multicolumn{6}{c}{Data Cube 1 -- DC1} \\
\toprule\bottomrule
& $\text{NRMSE}_{\bA}$ & $\text{NRMSE}_{\bbM}$ & $\text{SAM}_{\bbM}$ & $\text{NRMSE}_{\bY}$ & Time [s] \\ \midrule
\toprule	
FCLS	&	0.2854	&	---	&	---	&	0.0350	&	0.71	\\
PLMM	&	0.2604	&	0.1075	&	0.0440	&	\textbf{0.0007}	&	122.09	\\
ELMM	&	0.2554	&	0.1032 	&	0.0398 	&	0.0321 	&	8.82	\\
GLMM	&	0.2480	&	0.1036	&	0.0355	&	0.0235	&	23.74	\\
DeepGUn	&	\textbf{0.0566}	&	\textbf{0.0944}	&	\textbf{0.0233}	&	0.0448	&	75.20	\\
\bottomrule
\multicolumn{6}{c}{Data Cube 2 -- DC2}	\\
\toprule\bottomrule	
FCLS	&	0.1294	&	---	&	---	&	0.0393	&	0.38	\\
PLMM	&	0.1197 	&	0.0481 	&	0.0378 	&	0.0336 	&	41.31	\\
ELMM	&	0.1110 	&	0.0566 	&	0.0382 	&	0.0231 	&	20.25	\\
GLMM	&	0.1146	&	0.0534 	&	0.0367	&	\textbf{0.0226}	&	17.03	\\
DeepGUn	&	\textbf{0.0969}	&	\textbf{0.0463}	&	\textbf{0.0323}	&	0.0384	&	36.40	\\
\bottomrule
\multicolumn{6}{c}{Data Cube 3 -- DC3}	\\
\toprule\bottomrule	
FCLS	&	0.2606	&	---	&	---	&	0.0542	&	0.34	\\
PLMM	&	0.2028 	&	0.0928 	&	0.0385 	&	0.0302 	&	59.88	\\
ELMM	&	0.1997 	&	0.0640 	&	0.0188 	&	0.0238 	&	17.99	\\
GLMM	&	0.1841	&	0.0638	&	0.0185	&	\textbf{0.0226}	&	25.74	\\
DeepGUn	&	\textbf{0.1613}	&	\textbf{0.0600}	&	\textbf{0.0172}	&	0.0457	&	48.96	\\
\bottomrule
\multicolumn{6}{c}{Data Cube 4 -- DC4}	\\
\toprule\bottomrule	
FCLS	&	0.5109	&	---	&	---	&	0.1712	&	0.50	\\
PLMM	&	0.5066 	&	0.6245 	&	0.4874 	&	0.0320 	&	269.48	\\
ELMM	&	0.4385 	&	0.4712 	&	0.1451 	&	\textbf{0.0106} 	&	18.36	\\
GLMM	&	0.4371	&	0.4855	&	0.1972	&	0.0108	&	21.15	\\
DeepGUn	&	\textbf{0.2550}	&	\textbf{0.2918}	&	\textbf{0.0873}	&	0.1403	&	99.94	\\
\toprule		
\end{tabular}
\label{tab:results_synthData}
\end{table}


\begin{figure}[!htb] 
\centering
\includegraphics[height=0.45\textwidth, width=0.28\textwidth, angle=-90,origin=c]{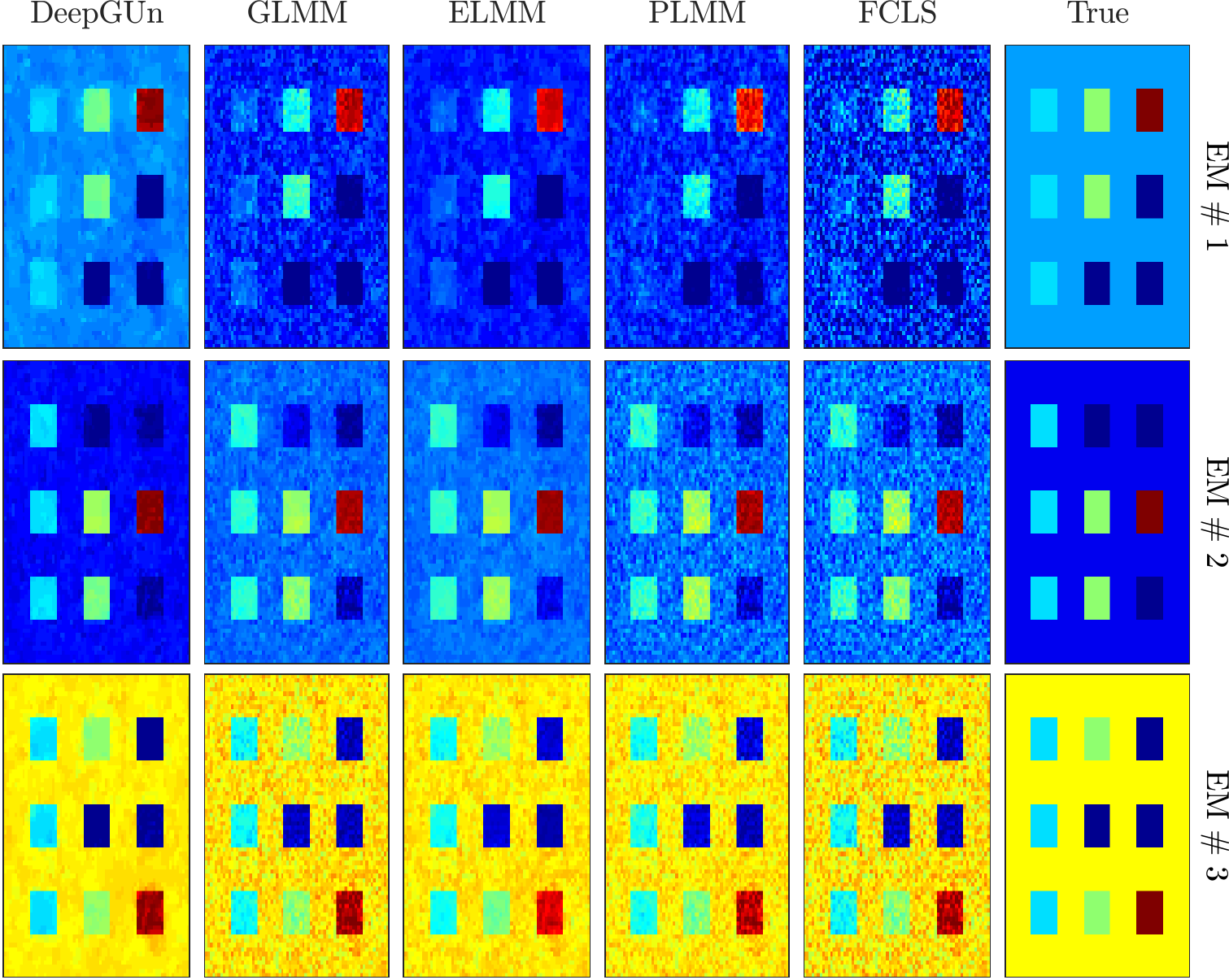} \\
\vspace{-1.4cm}
\includegraphics[height=0.45\textwidth, width=0.28\textwidth, angle=-90,origin=c]{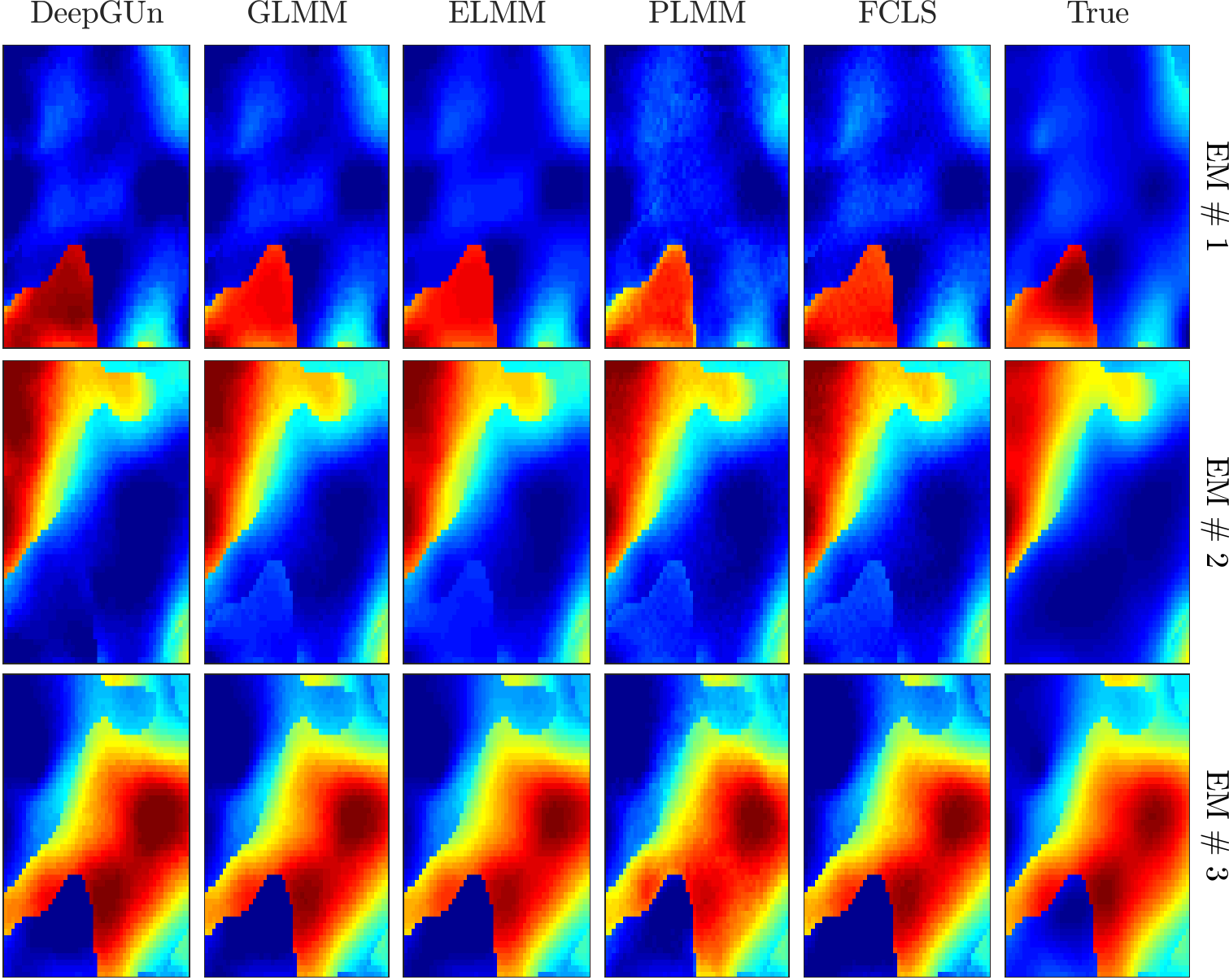}\\
\vspace{-1.4cm}
\includegraphics[height=0.45\textwidth, width=0.28\textwidth, angle=-90,origin=c]{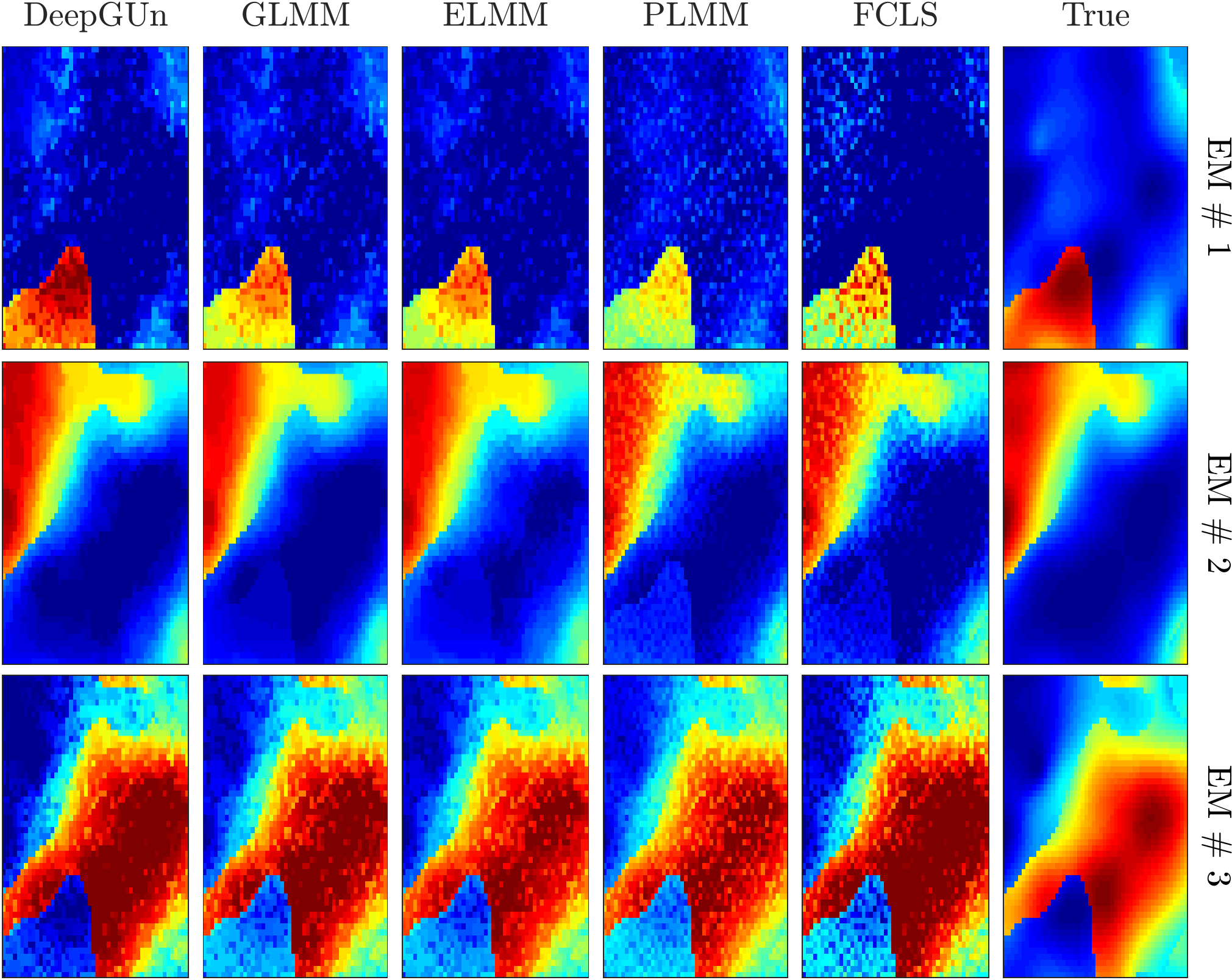}\\
\vspace{-1.4cm}
\includegraphics[height=0.45\textwidth, width=0.28\textwidth, angle=-90,origin=c]{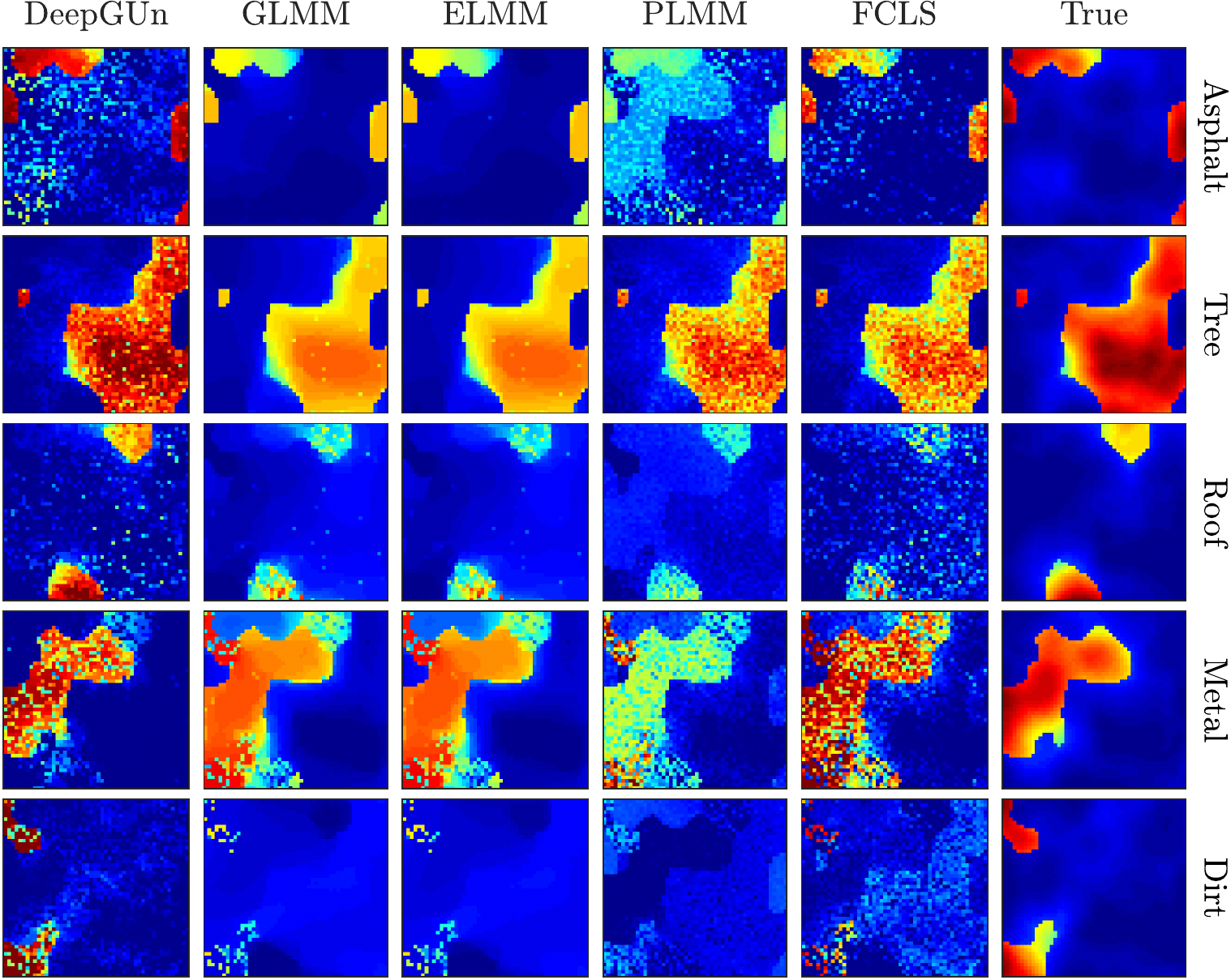}
\vspace{-1.4cm}
\caption{Abundance maps of DC1 (top), DC2 (middle up), DC3 (middle down) and DC4 (bottom) for all tested algorithms. Abundance values represented by colors ranging from blue ($a_k=0$) to red ($a_k=1$).}
\label{fig:ab_maps_synth}
\end{figure}


\subsection{Synthetic data}


To quantitatively compare the different algorithms, four synthetic datasets were created, namely Data Cubes~1--4 (DC1--DC4), with 70$\times$70 pixels (DC1) and 50$\times$50 pixels (DC2, DC3 and DC4). These datasets were built using three (DC1, DC2 and DC3) and five (DC4) 224-band EMs extracted from the USGS Spectral Library~\cite{clark2003imaging} and spatially correlated abundance maps, as depicted in the first row of Fig.~\ref{fig:ab_maps_synth}.

Spectral variability of the EMs was imposed using four different models. For the DC1 datacube, we adopted the variability model used in~\cite{Thouvenin_IEEE_TSP_2016_PLMM}, consisting of pixelwise multiplicative spectral factors given by random piecewise-linear functions. For DC2, the variability model of~\cite{imbiriba2018glmm} was used, consisting of band dependent scaling factors that varied smoothly in both the spatial and spectral dimensions. For DC3, we considered a simple model introduced in~\cite[Section~IV-A-1]{uezato2016novelBundlesExtractionClustering} to emulate errors in atmospheric compensation as a function of the viewing geometry given the direct and diffuse light on the scene, and the solar path transmittance. For datacube DC4, we used as endmembers pure pixels of five materials (asphalt, tree, roof, metal and dirt) which were manually extracted from a real hyperspectral image, thus depicting realistic spectral variability. White Gaussian noise was finally added to all datasets to yield a 30dB SNR.


The optimal parameters for each algorithm were selected by performing grid searches for each dataset. The ranges in which the parameters were searched were selected according to those discussed by the authors in the original publications. For the PLMM we searched for $\alpha$, $\beta$ and $\gamma$ in the ranges $[0.01,\,0.1,\,0.35,\, 0.7,\, 1.4,\, 25]$, $[10^{-9},\, 10^{-5},\, 10^{-4},\, 10^{-3}]$ and $[10^{-2},\,0.1,\,1,\,10,\,10^2]$, respectively. For both ELMM and GLMM, the parameters were selected among the following values: $\lambda_{S},\,\lambda_M \in [0.01,\, 0.1,\, 1,\, 5,\, 10,\, 50]$, $\lambda_{A} \in [0.001,\, 0.01,\, 0.05]$, and $\lambda_\psi,\,\lambda_{\bbPsi} \in [10^{-6},\, 10^{-3},\, 1,\,10^{3}]$. For the proposed DeepGUn algorithm, we fixed $\lambda_Z=0.1$ and selected $\lambda_A$ among the values $[0.005,\, 0.01,\,0.05]$.
For the proposed method, the sets of pure pixels for each EM $\mathcal{N}_{\mathcal{P},p}$ were constructed by selecting the~$100$ pixels closest to the reference materials~$\bM_0$.

The quantitative results are shown in Table~\ref{tab:results_synthData}, with the best results for each metric marked in bold. The proposed method clearly outperformed the competing algorithms in terms of $\text{NRMSE}_{\bA}$ for all four datasets. Qualitatively, the abundance maps provided by DeepGUn, displayed in Fig.~\ref{fig:ab_maps_synth}, are clearly much closer to the true abundance maps than those provided by the other methods. These are important results, as accuracy in abundance estimation is the main objective of SU.


For the EM reconstruction metrics $\text{NRMSE}_{\bbM}$ and $\text{SAM}_{\bbM}$, DeepGUn gave the best results for all data cubes. This indicates that the proposed endmember model used by DeepGUn allows for precise material identification from the observed hyperspectral scenes.

The reconstruction error $\text{NRMSE}_{\bY}$ of the DeepGUn algorithm was comparable to the FCLS and significantly larger than that of the GLMM. This is natural since the GLMM has more degrees of freedom. However, the connection between $\text{NRMSE}_{\bY}$ and the abundance reconstruction error is far from being direct, as can be seen in Table~\ref{tab:results_synthData}. 

The execution times, at the rightmost column of Table~\ref{tab:results_synthData}, indicate that the computational complexity of DeepGUn is somewhere between the complexities of GLMM and PLMM, the two major  competing algorithms. Hence, the DeepGUn method yielded superior SU performance, with easier parameter tuning, and at a reasonable computational cost.

\subsection{Influence of the latent dimension~$K$}

An important parameter in the design of the DeepGUn method is the dimensionality $K$ of the latent space of the generative endmember models $\mathcal{G}_i$, $i=1,\ldots,P$. The individual dimensions of the latent space are used to represent changes in the endmember signatures due to spectral variability. Since in a given scene the endmembers are likely to be affected only by a small number of effects (hence the hypothesis that they are supported at a low-dimensional manifold),~$K$ should be small in order to avoid introducing spurious effects and increasing the computational complexity of the solution.

\begin{figure}[htb!]
    \centering
    \includegraphics[width=0.47\linewidth]{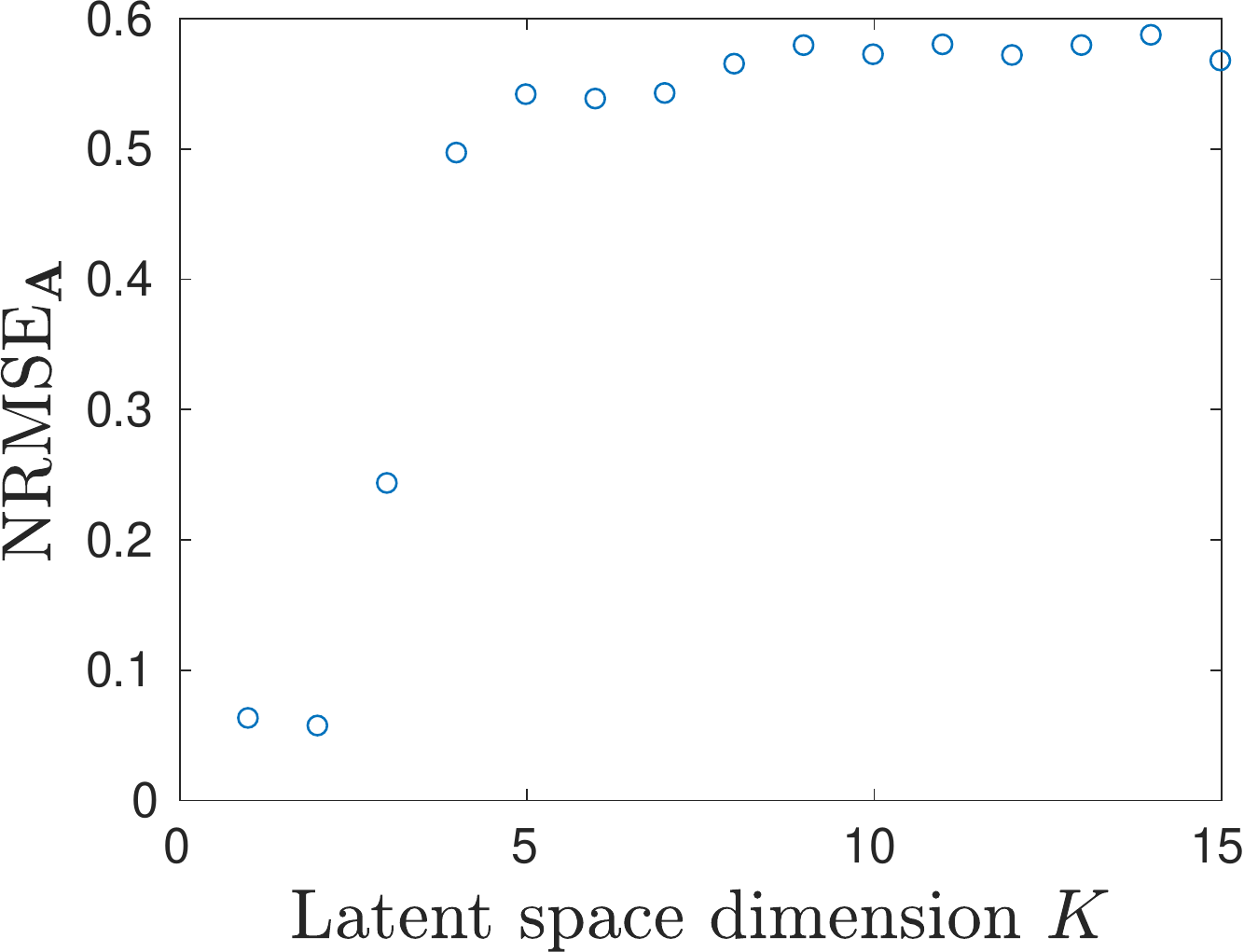} \,
    \includegraphics[width=0.47\linewidth]{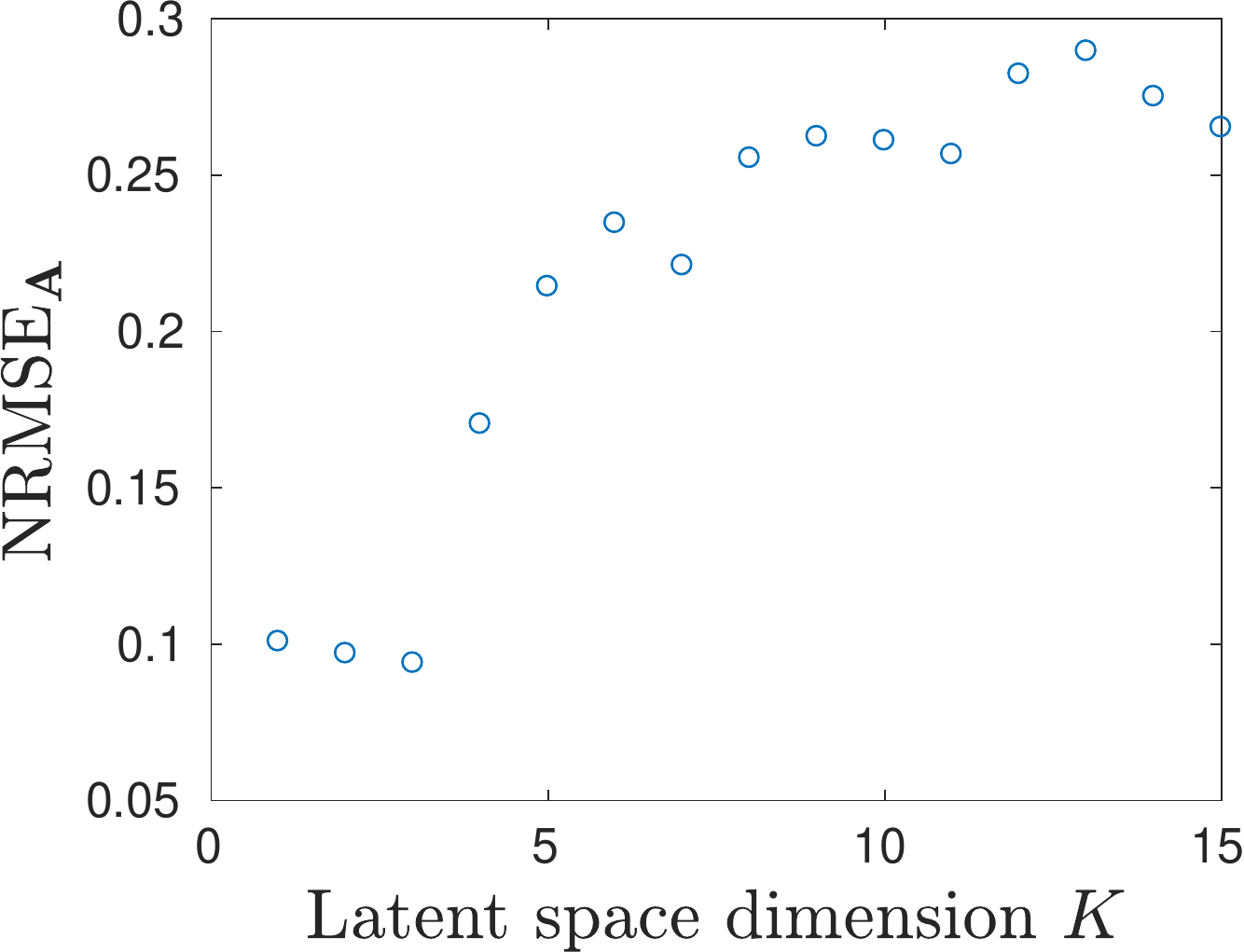}
    \caption{Abundance NRMSE as a function of the latent space dimension~$K$ for datacubes DC1 (left) and DC2 (right).}
    \label{fig:err_vs_latent_dim_K}
\end{figure}

To illustrate this, we performed a simulation with DeepGUn where we varied the dimensionality of the latent space~$K$ and measured the normalized abundance reconstruction error~$\text{NRMSE}_{\bA}$. For this, we considered the data cubes DC1 and DC2 described above in Section~VII-A. The results are depicted in Figure~\ref{fig:err_vs_latent_dim_K}, and show that the abundance estimation error tends to increase with~$K$. It can also be seen that there is a sharper increase in~$\text{NRMSE}_{\bA}$ for DC1 when compared to DC2. This is likely due to the fact that the dataset DC2 uses a more complex model to generate endmember variability. This indicates that the selection of a small value for~$K$ is important to obtain good unmixing results.

\begin{table}[htb!]
\caption{Simulation results using real data.}
\vspace{-0.2cm}
\centering
\renewcommand{\arraystretch}{1.1}
\begin{tabular}{
@{\hspace{0.3\tabcolsep}} l 
@{\hspace{0.75\tabcolsep}} |
@{\hspace{0.75\tabcolsep}} c 
@{\hspace{0.75\tabcolsep}} c
@{\hspace{0.75\tabcolsep}} |
@{\hspace{0.75\tabcolsep}} c 
@{\hspace{0.75\tabcolsep}} c
@{\hspace{0.75\tabcolsep}} |
@{\hspace{0.75\tabcolsep}} c 
@{\hspace{0.75\tabcolsep}} c
@{\hspace{0.5\tabcolsep}}}
\bottomrule
& \multicolumn{2}{@{\hspace{-1.8\tabcolsep}}c@{\hspace{-1.5\tabcolsep}}|@{\hspace{0.75\tabcolsep}}}{Houston HI} 
& \multicolumn{2}{@{\hspace{-0.5\tabcolsep}}c@{\hspace{-0.1\tabcolsep}}|@{\hspace{0.75\tabcolsep}}}{Samson HI} 
& \multicolumn{2}{@{\hspace{-1\tabcolsep}}c@{\hspace{0.5\tabcolsep}}}{Jasper Ridge HI} \\
\midrule
& $\scalemath{0.85}{\text{NRMSE}_{\bY}}$ 
& $\scalemath{0.95}{\text{Time [s]}}$ 
& $\scalemath{0.85}{\text{NRMSE}_{\bY}}$ 
& $\scalemath{0.95}{\text{Time [s]}}$ 
& $\scalemath{0.85}{\text{NRMSE}_{\bY}}$ 
& $\scalemath{0.95}{\text{Time [s]}}$  \\ 
\toprule
FCLS	&	0.2470	&	2.56	&	0.0545	&	1.38	&	0.2057	&	1.52	\\
PLMM	&	0.0713	&	663.25	&	0.0239	&	103.84	&	0.0553	&	220.84	\\
ELMM	&	0.0171	&	38.30	&	0.0119	&	14.76	&	0.0278	&	27.08	\\
GLMM	&	\textbf{0.0016}	&	48.53	&	\textbf{0.0006}	&	46.69	&	\textbf{0.0019}	&	86.33	\\
DeepGUn	&	0.2355	&	259.61	&	0.0862	&	121.88	&	0.1094	&	209.64	\\
\toprule
\end{tabular}
\label{tab:results_realData}
\end{table}

\begin{figure}[htb]
\centering
\includegraphics[width=\linewidth]{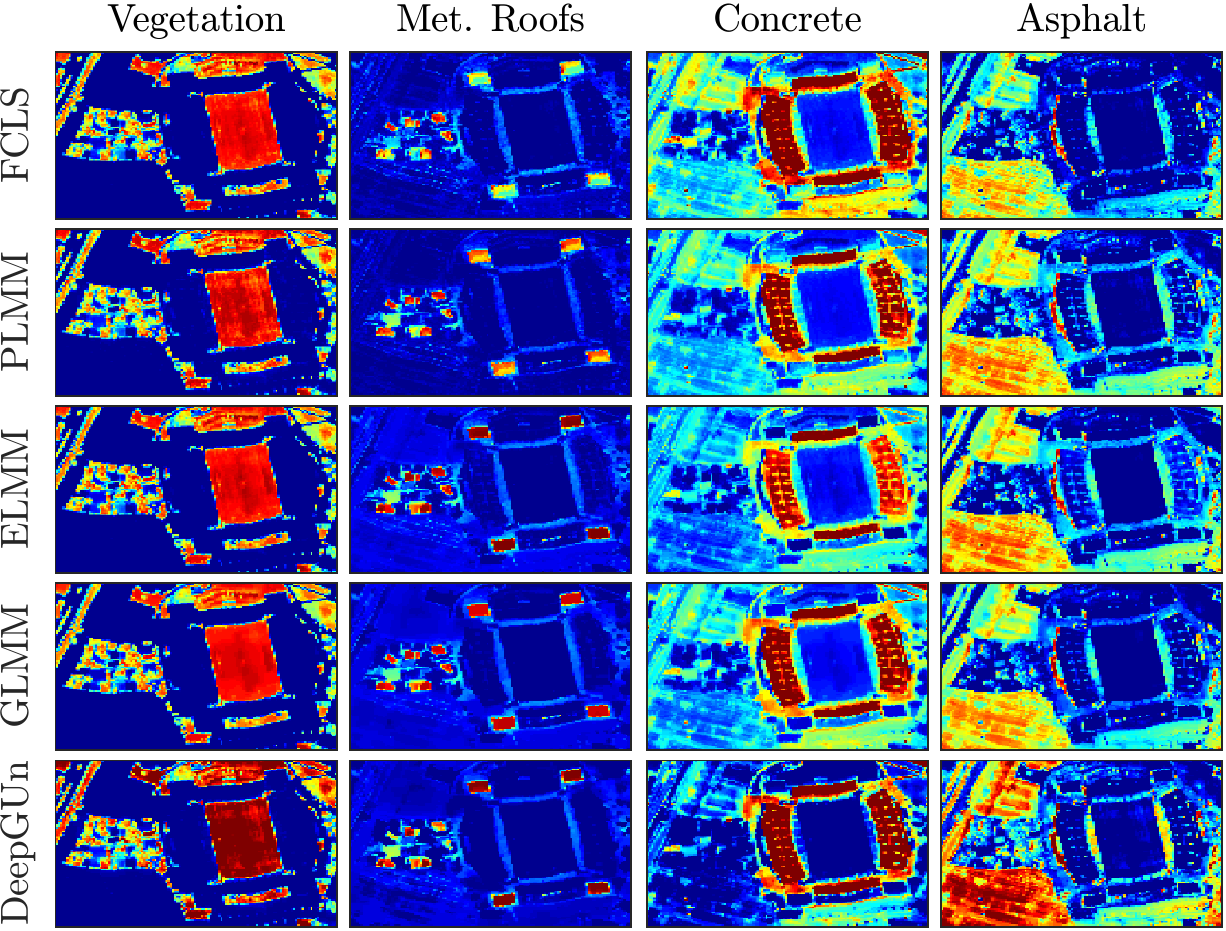}
\caption{Abundance maps of the Houston dataset for all tested algorithms. Abundance values represented by colors ranging from blue ($\alpha_k = 0$) to red ($\alpha_k = 1$).}\label{fig:ab_maps_houston}
\end{figure}


\begin{figure}[htb]
\centering
\includegraphics[width=0.8\linewidth]{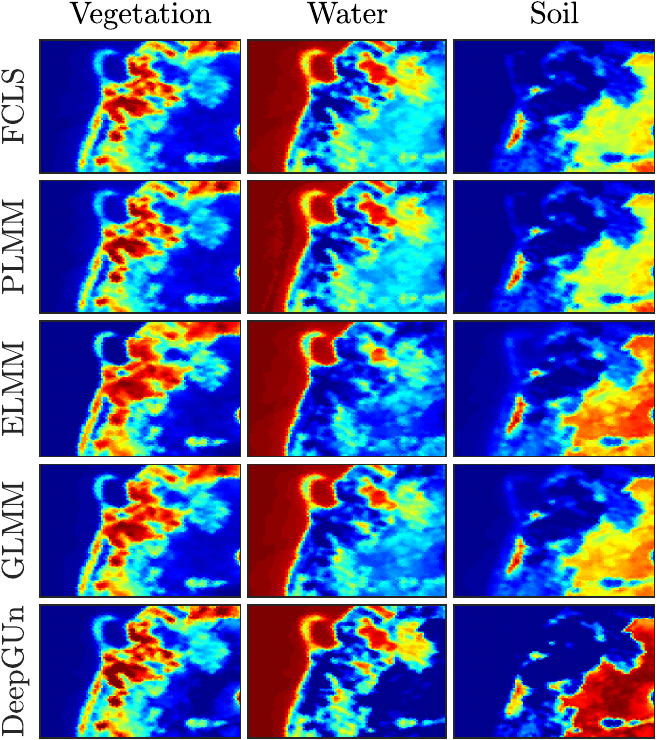}
\caption{Abundance maps of the Samson dataset for all tested algorithms. Abundance values represented by colors ranging from blue ($\alpha_k = 0$) to red ($\alpha_k = 1$).}\label{fig:ab_maps_samson}
\end{figure}
\begin{figure}[htb]
\centering
\includegraphics[width=0.9\linewidth]{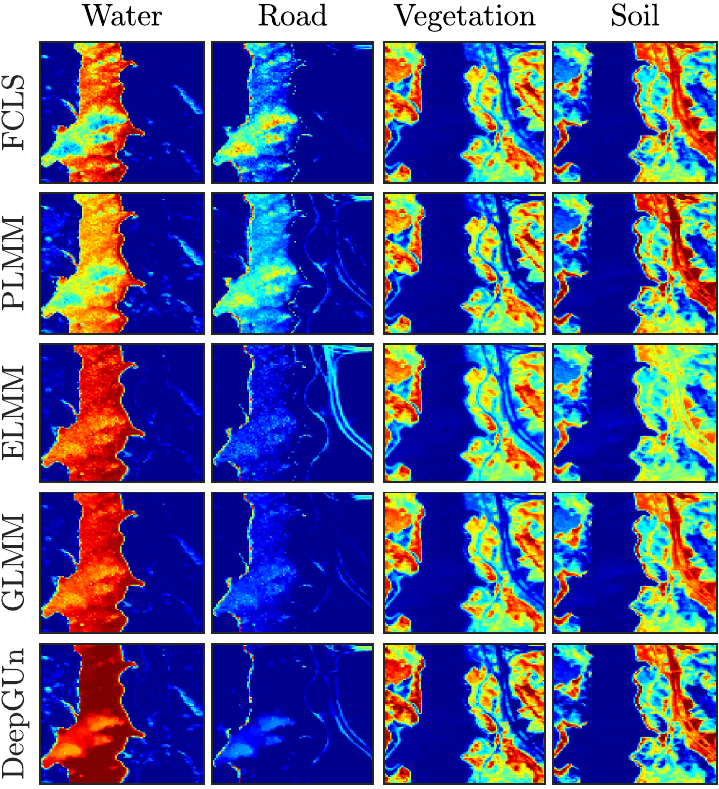}
\caption{Abundance maps of the Jasper Ridge dataset for all tested algorithms. Abundance values represented by colors ranging from blue ($\alpha_k = 0$) to red ($\alpha_k = 1$).}\label{fig:ab_maps_jasper}
\end{figure}

\subsection{Real data}
We considered the Houston, Samson and the Jasper Ridge datasets for the simulations with real data. These datasets were captured by the AVIRIS instrument, and originally had~224 bands. The spectral bands corresponding to water absorption and low SNR regions were removed, resulting in~188 bands for the Houston image,~156 bands for the Samson image and~198 bands for the Jasper Ridge image.
Previous studies indicate that the Houston HI has four predominant EMs~\cite{drumetz2016blindUnmixingELMM}, while the Samson and Jasper Ridge HIs are known to have three and four EMs, respectively~\cite{Borsoi2017_multiscale}.

The reconstructed abundance maps for both datasets and all algorithms are shown in Figs.~\ref{fig:ab_maps_houston},~\ref{fig:ab_maps_samson} and~\ref{fig:ab_maps_jasper}.
For the Houston dataset, the last row of Fig.~\ref{fig:ab_maps_houston} shows that the abundance maps provided by the DeepGUn method better evidence the strong vegetation and concrete abundances at the stadium field and stands, respectively, as well as the stronger asphalt abundances in the parking lot.
%
%
For the Samson and Jasper Ridge images, a clear performance improvement can be seen for the DeepGUn algorithm. Note, for instance, a smaller confusion between the Water and Soil EMs in the Samson HI when compared to the other methods. Similarly, for the Jasper Ridge HI, the DeepGUn method leads to considerably stronger Water abundances in the region containing the river. Moreover, although the ELMM provided a better estimation of the road in the scene when compared to the remaining methods, it also resulted in a greater confusion between the Vegetation and Soil EMs, especially in the right part of the scene.

The quantitative results for all algorithms and datasets are shown in Table~\ref{tab:results_realData}. Since the correct abundance values (the ground truth) are not available for most real images, the reconstruction error $\text{NRMSE}_{\bY}$ has been used as a sort of quality verification. As was the case for the synthetic data, the DeepGUn reconstruction errors are higher than those yielded by other methods that address spectral variability. However, the reconstruction error is definitely not a good performance measure for abundance estimation in real images, which is the main objective of unmixing algorithms. The higher reconstruction errors of DeepGUn in this case are just due the fact that DeepGUn has much fewer degrees of freedom than the ELMM, PLMM and GLMM algorithms. In fact, the DeepGUn has only \mbox{$K\times P$} degrees of freedom for each pixel, which is comparable to the FCLS ($P$) much smaller than the ELMM, GLMM and PLMM methods (\mbox{$>L\times P$}). Although this means that the ELMM, GLMM and PLMM can achieve arbitrarily small reconstruction errors~$\text{NRMSE}_{\bY}$, this is not necessarily reflected as good abundance estimation results.
The execution times of the proposed DeepGUn method were again comparable to those of the other algorithms addressing spectral variability, which indicates that it scales well with larger image sizes.

\section{Conclusions} \label{sec:conclusions}


In this paper, a deep generative EM model was proposed to address spectral variability in SU of HIs. Instead of relying on user-defined parametric EM models which have shown to be very hard to estimate in practical scenes, the proposed methodology leveraged the generalization capability of deep neural networks to accurately model EM spectra while still maintaining a strong connection to the physical mixing process. 
A deep generative model for each EM was trained prior to unmixing by using pure pixel information extracted directly from the observed HI, which allowed for an unsupervised formulation.
The proposed EM model was then applied to solve the SU problem, which was cast as the estimation of the low-dimensional representations of the EMs in the latent space of the deep generative models and their corresponding fractional abundances, for each pixel.
The resulting DeepGUn algorithm presented excellent performance despite the simple strategy used for selecting the training data for learning the generative model.
%
Simulations using synthetic and real data indicate that the proposed method can lead to significant improvements in abundance estimation accuracy.

\bibliographystyle{IEEEtran}
\bibliography{references_revpaper,references_DpGen,ourpapers}

\vspace{-1cm}
\begin{IEEEbiographynophoto}{Ricardo Augusto Borsoi (S'18)} 
received the MSc degree in electrical engineering from Federal University of Santa Catarina (UFSC), Florian\'opolis, Brazil, in 2016. He is currently working towards his doctoral degree at Universit\'e C\^ote d'Azur (OCA) and at UFSC. His research interests include image processing, tensor decomposition, and hyperspectral image analysis.
\end{IEEEbiographynophoto}

\vspace{-1cm}
\begin{IEEEbiographynophoto}{Tales Imbiriba (S'14, M'17)}   
received his Doctorate degree from the Department of Electrical Engineering (DEE) of the Federal University of Santa Catarina (UFSC), Florian\'opolis, Brazil, in 2016. He served as a Postdoctoral Researcher at the DEE--UFSC and is currently a Postdoctoral Researcher at the ECE dept. of the Northeastern University, Boston, MA, USA. 
His research interests include audio and image processing, pattern recognition, kernel methods, adaptive filtering, and Bayesian Inference.
\end{IEEEbiographynophoto}

\vspace{-1cm}
\begin{IEEEbiographynophoto}{Jos\'e Carlos M. Bermudez (S'78,M'85,SM'02)}
received the B.E.E. degree from the Federal University of Rio de Janeiro (UFRJ), Rio de Janeiro, Brazil, the M.Sc. degree in electrical engineering from COPPE/UFRJ, and the Ph.D. degree in electrical engineering from Concordia University, Montreal, Canada, in 1978, 1981, and 1985, respectively.
  He joined the Department of Electrical Engineering, Federal University of Santa Catarina (UFSC), Florianopolis, Brazil, in 1985. He is currently a Professor of Electrical Engineering at UFSC and a Professor at Catholic University of Pelotas (UCPel), Pelotas, Brazil. He has held the position of Visiting Researcher several times for periods of one month at the Institut National Polytechnique de Toulouse, France, and at Université Nice Sophia-Antipolis, France. He spent sabbatical years at the Department of Electrical Engineering and Computer Science, University of California, Irvine (UCI), USA, in 1994, and at the Institut National Polytechnique de Toulouse, France, in 2012. 
  His recent research interests are in statistical signal processing, including linear and nonlinear adaptive filtering, image processing, hyperspectral image processing and machine learning.
  Prof. Bermudez served as an Associate Editor of the IEEE TRANSACTIONS ON SIGNAL PROCESSING in the area of adaptive filtering from 1994 to 1996 and from 1999 to 2001. He also served as an Associate Editor of the EURASIP Journal of Advances on Signal Processing from 2006 to 2010, and as a Senior Area Editor of the IEEE TRANSACTIONS ON SIGNAL PROCESSING from 2015 to 2019.  He is the Chair of the Signal Processing Theory and Methods Technical Committee of the IEEE Signal Processing Society. Prof. Bermudez is a Senior Member of the IEEE.


\end{IEEEbiographynophoto}

\end{document}